\pdfoutput=1

\documentclass[11pt]{article}

\usepackage[final]{acl}
\usepackage{enumitem}
\usepackage{times}
\usepackage{latexsym}

\usepackage[T1]{fontenc}

\usepackage[utf8]{inputenc}

\usepackage{microtype}

\usepackage{inconsolata}

\usepackage{inconsolata}
\usepackage{comment}
\usepackage{graphicx}
\usepackage{multirow}
\usepackage{array}
\usepackage{wrapfig}
\usepackage{amsmath}
\usepackage{makecell}
\usepackage{tabularx}
\usepackage{booktabs}
\usepackage{cleveref}
\usepackage{amsfonts}
\usepackage[table]{xcolor}
\usepackage{wrapfig}
\usepackage{tcolorbox}
\tcbuselibrary{skins}
\usepackage{pifont}
\newcommand{\cmark}{\ding{51}}  
\newcommand{\xmark}{\ding{55}}
\usepackage{listings}
\usepackage{xcolor}

\definecolor{lightgray}{gray}{0.95}

\usepackage{blindtext}

\usepackage{color, colortbl}
\definecolor{Yellow}{rgb}{1.0,0.98,0.8}

\title{MINDS: A Cross-cultural Dialogue Corpus for \\Social Norm Classification and Adherence Detection}

\author{
 \textbf{Pritish Sahu\thanks{Equal contribution}},
 \textbf{Anirudh Som\footnotemark[1]},
 \textbf{Dimitra Vergyri},
 \textbf{Ajay Divakaran}
\\
 SRI International
\\
\small{
\texttt{\{pritish.sahu, contact.anirudh.som\}@gmail.com, \{dimitra.vergyri, ajay.divakaran\}@sri.com}
}
}

\begin{document}
\maketitle

\begin{abstract}
Social norms are implicit, culturally grounded expectations that guide interpersonal communication. Unlike factual commonsense, norm reasoning is subjective, context-dependent, and varies across cultures, posing challenges for computational models. Prior works provide valuable normative annotations but mostly target isolated utterances or synthetic dialogues, limiting their ability to capture the fluid, multi-turn nature of real-world conversations. In this work, we present \textbf{Norm-RAG}, a retrieval-augmented, agentic framework for nuanced social norm inference in multi-turn dialogues. Norm-RAG models utterance-level attributes including communicative intent, speaker roles, interpersonal framing, and linguistic cues and grounds them in structured normative documentation retrieved via a novel Semantic Chunking approach. This enables interpretable and context-aware reasoning about norm adherence and violation across multilingual dialogues. We further introduce \textbf{MINDS} (Multilingual Interactions with Norm-Driven Speech), a bilingual dataset comprising 31 multi-turn Mandarin-English and Spanish-English conversations. Each turn is annotated for norm category and adherence status using multi-annotator consensus, reflecting cross-cultural and realistic norm expression. Our experiments show that Norm-RAG improves norm detection and generalization, demonstrates improved performance for culturally adaptive and socially intelligent dialogue systems.
\end{abstract}

\section{Introduction}\label{section_introduction}
Social norms are culturally embedded, often implicit expectations that shape how individuals interact in society, especially in interpersonal dialogues. These norms whether, behavioral conventions, moral obligations, or expectations of politeness guide acceptable conduct and influence both verbal and non-verbal communication \cite{sherif1936psychology, haidt2012righteous, schwartz2012refining}. 
As norms can vary significantly across cultures \cite{triandis1994culture, arieli1964cultural}, modeling them in computational systems demands reasoning beyond the literal meaning of utterances. Unlike factual commonsense, social norm reasoning involves subjectivity, context, and cultural nuances, making it significantly more ambiguous and under-determined. 

Recent works, however, have made strides in codifying normative knowledge through structured resources such as SocialChem \cite{forbes2020social}, NormSage \cite{fung2022normsage}, NormDial \cite{li2023normdial}, SocialDial \cite{zhan2023socialdial} and RENOVI \cite{zhan2024renovi}, which annotate social norms in descriptive scenarios or conversations. These datasets highlight the importance of understanding values, intents, justifications, and social expectations. However, they fall short in modeling the dynamic, multi-turn nature of conversations where shifts in intent, emotional alignment, or interpersonal sensitivity can dramatically affect norm interpretation. For example, identifying whether a speaker's disagreement is socially acceptable may depend on factors such as power dynamics, tone, or context established across earlier turns. Existing resources often lack fine-grained annotations for cues, interpersonal relationships, or latent intentions which are critical signals for robust social norm understanding.

To address these limitations we propose \textbf{Norm-RAG}, a retrieval-augmented generation based agentic framework for social norm inference in dialogue. It models utterance-level features such as intent, cue, interest, and role interplay, enabling a more nuanced detection of implicit and culturally sensitive social norms in natural conversation. Furthermore, it decomposes norm behavior into the following pragmatically grounded attributes to enhance the interpretability and accuracy of norm detection - \emph{Communicative Intent}, \emph{Interpersonal Framing}, \emph{Linguistic Features}, and \emph{Contextual Triggers and Constraints}. By integrating these structured representations with LLM-based reasoning and dynamic retrieval of relevant norm documentation, we support both norm classification and norm adherence/violation assessment in multilingual conversation settings.

We also release \textbf{MINDS}, short for \textbf{M}ultilingual \textbf{I}nteractions with \textbf{N}orm-\textbf{D}riven \textbf{S}peech, a novel bilingual dataset consisting of 31 annotated multi-turn dialogue sessions across Mandarin-English and Spanish-English pairs. Each turn is labeled for norm category and adherence/violation status labels, with multi-annotator consensus to ensure quality and consistency. Unlike prior datasets that rely on static prompts or synthetic conversations, ours captures natural, two-party interactions, enabling more realistic modeling of culturally embedded norm violations and their detection.

\noindent Our key contributions are summarized as follows:
\begin{itemize}[itemsep=1pt, topsep=2pt]
    \item We developed \textbf{Norm-RAG}, a novel agentic architecture that models social norms through retrieval-augmented generation, leveraging feedback from prior utterances and structured dialog context for turn-level inference.
    \item We introduce the \textbf{MINDS corpus}, a bilingual, multi-annotated dialogue dataset covering Spanish-English and Mandarin-English conversations, annotated for social norm type and adherence/violation status, reflecting cross-cultural, real-world interactions.
    \item We formulate norm classification using four interpretable, latent dimensions: \emph{Intent}, \emph{Framing}, \emph{Linguistic Features}, and \emph{Constraints}, moving beyond surface-level cue detection.
    \item We present a novel \textbf{Semantic Chunking} technique for norm document retrieval, replacing heuristic keyword matching with context-aware semantic segmentation to accurately extract applicable normative guidance.
    \item We benchmark and analyze retrieval-based and generative approaches across various configurations, demonstrating improved norm detection performance and generalizability.
\end{itemize}
\section{Related Work}\label{section_related-work}
Early efforts in modeling social norms computationally have centered around static textual contexts \cite{ziems2023normbank, sap2019atomic, rashkin2018event2mind, emelin2020moral, jiang2021can, kim2022prosocialdialog, gu2021dream, ziems2022moral, ch2023sociocultural}. Social Chemistry 101 \cite{forbes2020social} introduced Rules-of-Thumb (RoTs) which are defined as free-text normative statements tied to situational prompts annotated with categorical labels capturing legality, moral foundations, and cultural expectations. While this large-scale resource enabled pretraining norm-aware language models, its static, monologic format restricts application to real-time dialogic settings. To overcome the lack of interactive context, NormSage \cite{fung2022normsage} proposed a zero-shot prompting method for extracting culture-specific norms from dialogues across languages using GPT-3, creating the NormsKB knowledge base. Though NormSage allows dynamic norm discovery and cross-cultural applicability, it lacks annotations for turn-level adherence or violation, thereby limiting its utility in norm-tracking tasks. 

NormDial \cite{li2023normdial} advanced the field by annotating each dialogue turn with adherence, violation, or irrelevance labels, using human-in-the-loop generation of synthetic conversations grounded in American and Chinese norms. However, its reliance on synthetic data and predefined norm templates limits its coverage of spontaneous and organically evolving dialogues. SocialDial \cite{zhan2023socialdial} is a large-scale, monocultural resource centered on a Chinese ontology of social norms (5 categories, 14 subcategories). Its evaluation tasks involve predicting dialogue-level social factors (\emph{e.g.}, distance, relation, location, formality, topic) and detecting norm violations within Chinese cultural contexts. In contrast, our work introduces a cross-cultural, bilingual dataset spanning Mandarin–English and Spanish–English conversations derived from real conversational data, annotated not only for norm adherence but also for underlying speaker-level features such as intent, cue, and interpersonal alignment. Furthermore, we propose a dual-task framework that performs latent norm discovery alongside turn-level adherence classification, without assuming access to predefined norm statements. This approach offers a more holistic and dynamic modeling of social norms in conversation, addressing key limitations in prior datasets and moving closer to deployable, socially intelligent dialog systems.
\begin{figure*}[t]
    \centering
    \includegraphics[width=\linewidth]{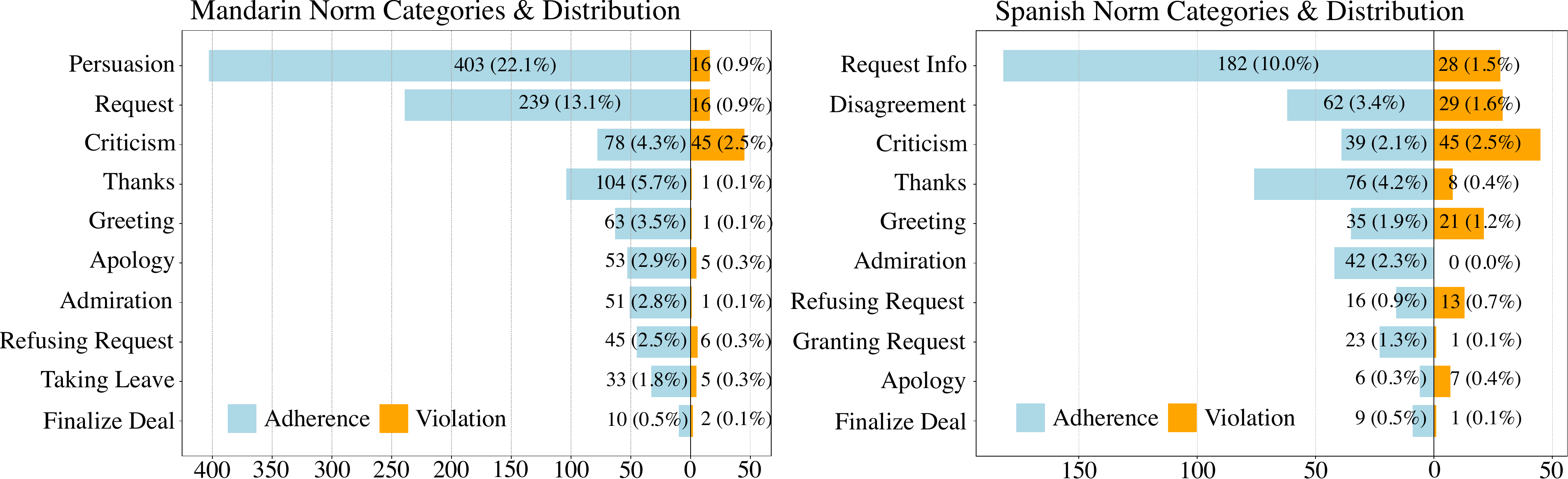}
    \caption{Distribution of annotated social norms across languages, norm categories and status (adherence/violation) labels. Numbers in parenthesis indicate sample percentage contribution in the database.}
    \label{fig:dist}
    \vspace{-0.15in}
\end{figure*}
\section{MINDS Corpus}\label{section_dataset}
\vspace{-0.4em}
Unlike static rule-sets, norms are context-sensitive, adaptive, and frequently nuanced by linguistic, interpersonal, and situational cues. Existing datasets have made significant strides in curating large-scale corpora for norm recognition and classification. However, each of these datasets presents notable limitations as described in Section \ref{section_related-work}. 
In contrast, the proposed MINDS corpus is curated from real bilingual conversational sessions with rich cultural and linguistic grounding. Below we outline the data collection and annotation methodology.

\subsection{Data Collection}\label{sub-sec_collection}
Each of the 31 sessions features a two-person, multi-turn dialogue between a foreign language expert and an English language expert. During the interaction, the foreign language expert communicates solely in the foreign language (either Mandarin or Spanish), while the English expert responds entirely in English. This structure was designed to simulate real-world bilingual communication scenarios such as interpreter training, second-language learning environments, or multilingual human-computer interactions.

The dataset is evenly balanced between the two language groups, where 16 sessions involve Mandarin speakers and 15 involve Spanish speakers. The Mandarin subset includes 12 unique speakers, with each unique speaker pair appearing only once. Similarly, 22 unique Spanish speakers were recruited with the same one-time unique pair participation constraint. This design ensures speaker diversity and eliminates redundancy, which is critical for fair evaluation in speaker-independent modeling tasks. The dialogues reflect a wide range of interpersonal dynamics, accents, and conversational styles. By enforcing the one-language-per-speaker rule, the setup captures code-switching boundaries, implicit translation patterns, and culturally specific communication strategies, making this dataset well-suited for multilingual NLP tasks.

\subsection{Annotation Protocol}\label{sub-sec_annotation}
Each session in the dataset was independently annotated by multiple human raters. The annotators were tasked with labeling individual utterances within the dialogue, assigning a social norm and a status (e.g., adherence or violation of the social norm). These judgments reflect the pragmatic and sociolinguistic interpretation of the utterances in context, requiring annotators to consider intent, tone, and interpersonal dynamics. The annotation process was structured such that no single annotator was responsible for all sessions in a given language group. Instead, annotators were selectively assigned to sessions, resulting in partial coverage across sessions and users. For instance, within the Mandarin subset, 5 unique annotators participated, with each annotating a subset of the 16 sessions. Similarly, 6 annotators contributed to the Spanish subset, with varied levels of session coverage. No annotator covered all sessions, a deliberate design choice to maintain annotator diversity and avoid individual annotator bias dominating the evaluation outcomes.

Each turn in a session was labeled by at least one annotator, and many were reviewed by two or more annotators, enabling inter-rater reliability (IRR) analysis. In cases where multiple annotators reviewed the same turn, the agreement was quantified using Cohen’s Kappa across combinations of norm and status annotations. Several sessions show high IRR, with only minor annotation discrepancies in labeling across turns, indicating high consistency where overlap existed. This further supports the reliability of the annotations in modeling and evaluation tasks. By leveraging a multi-rater framework and ensuring speaker-annotator diversity, this annotation protocol supports robust downstream applications such as norm adherence classification, multilingual dialogue modeling, and socially intelligent agent development. The combination of linguistic variation, dialogue realism, and reliable annotation makes this dataset a valuable benchmark for cross-cultural and multilingual AI systems.

\begin{figure}[ht!]
    \centering
    \includegraphics[width=\linewidth]{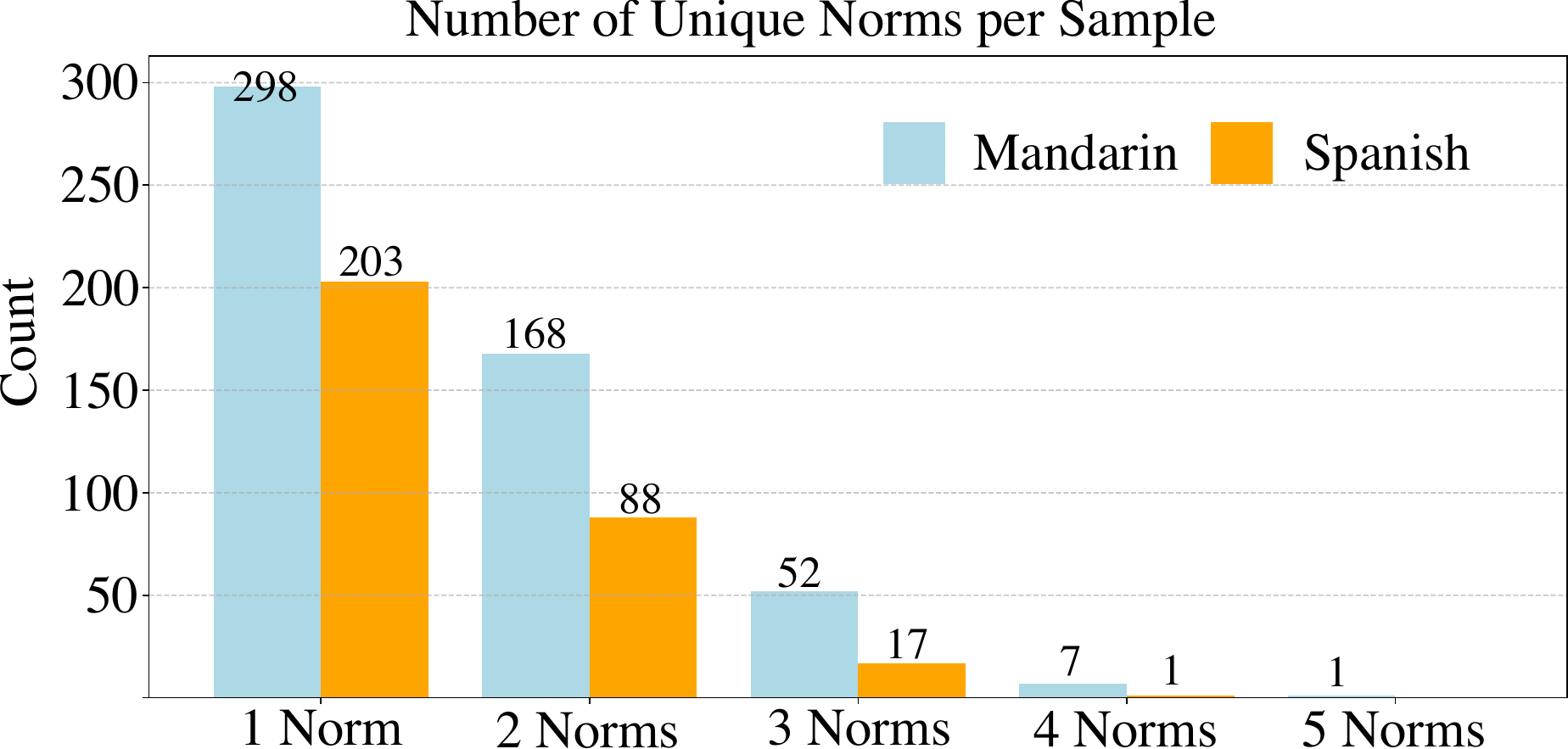}
    \caption{Frequency of norm categories per conversation sample.}
    \label{fig:norm_sample}
    \vspace{-0.25in}
\end{figure}

\subsection{Insights}\label{sub-sec_insights}

As illustrated in Figure \ref{fig:dist}, Mandarin-English dialogues show a strong skew toward Persuasion and Request categories, suggesting a focus on negotiation and directive strategies. In contrast, Spanish-English interactions are dominated by Request for Information and Thanks, reflecting a more transactional or expressive discourse style. These trends highlight language-specific tendencies in conversational norms. The relative frequency of Disagreement and Criticism in Spanish, compared to Persuasion in Mandarin, points to cultural differences in how interpersonal boundaries and assertiveness are navigated. Such patterns underscore the need for culturally adaptive modeling of normative behavior. Violations are infrequent overall but consistently concentrated in sensitive categories like Criticism and Disagreement across both languages. These acts, though less common, are more likely to deviate from normative expectations, indicating higher pragmatic risk.

Most utterances reflect norm adherence, suggesting it is the default mode of interaction. However, norm-sensitive acts such as refusals or criticisms, even when rare, carry higher likelihoods of violation highlighting the asymmetry in norm observance across speech acts. The dataset captures a broad spectrum of social norms with balanced coverage of adherence and violation, across two linguistically and culturally distinct bilingual contexts. This makes it well-suited for cross-cultural norm modeling and the development of socially aware dialogue systems. Figure \ref{fig:norm_sample} illustrates the distribution of the number of unique norm categories annotated per conversation sample, comparing Mandarin-English and Spanish-English dialogues. Most samples exhibit a single norm type, with 298 Mandarin and 203 Spanish samples falling into this category. As the number of distinct norms per sample increases, the counts drop sharply, indicating that multi-norm interactions are less common. Nevertheless, a notable portion of conversations particularly in Mandarin contain two or more co-occurring norms, highlighting the normative complexity present in real-world dialogue. This distribution supports the need for models capable of handling multi-label norm classification in conversational contexts.


\section{Approach}\label{section_approach}
\begin{figure*}[!t]
    \begin{center}
        \includegraphics[width=0.99\linewidth]{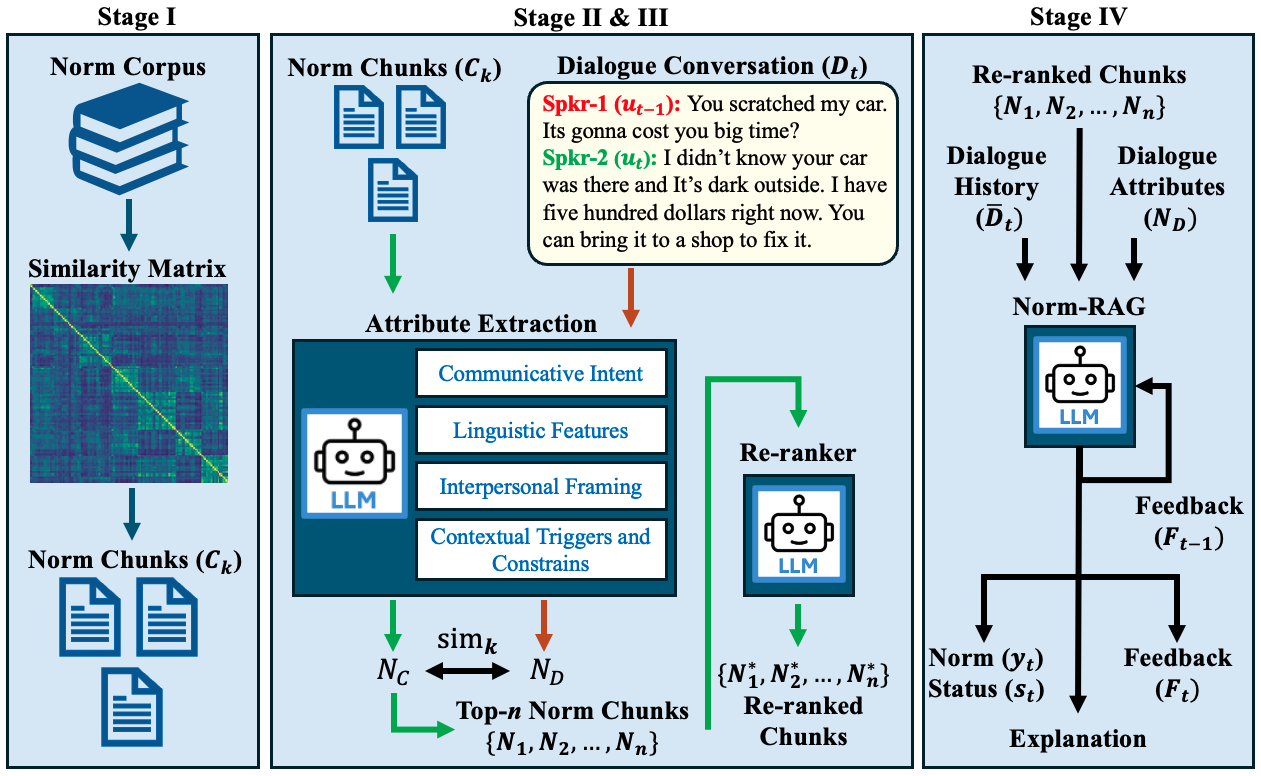}
    \end{center}
    \caption{Illustration of the different stages of our proposed framework. Stage I represents semantic clustering for norm chunking. Stage II \& III illustrates the structured norm attribute extraction, semantic norm retrieval and re-ranking modules. Finally, Stage IV shows the dialogue-aware norm classifier within the \textbf{Norm-RAG} system.}
    \label{fig:model}
    \vspace{-4mm}
\end{figure*}

Here, we describe our Retrieval-Augmented Generation framework, \textbf{Norm-RAG}, designed for social norm classification and adherence detection in dialogues. Unlike prior works that either retrieve from static corpora \cite{forbes2020social} or synthesize templated dialogue \cite{li2023normdial}, our method targets real-time, evolving conversational data. The central insight of \textbf{Norm-RAG} is to represent norms through 
pragmatic, multidimensional structures, enabling semantic and social interpretability beyond surface-level content. Our framework consists of four stages as illustrated in Figure \ref{fig:model}: (1) \textit{Semantic Clustering for Norm Chunking}, (2) \textit{Structured Norm Attribute Extraction}, (3) \textit{Semantic Norm Retrieval and Re-ranking}, (4) \textit{Dialogue-Aware Norm Classification}. Refer to Appendix \ref{sec:prompt} for details about the different prompt templates used throughout the entire process.

\subsection{Semantic Clustering for Norm Chunking}

To structure the norm corpus for retrieval and alignment, we adopt a block-diagonal clustering approach guided by semantic similarity between norm sentences. Inspired by prior work on block-diagonal clustering structures~\cite{xing2024block}, we utilize a similarity-based heuristic grounded in the observation that semantically coherent norm definitions exhibit localized similarity in embedding space.

Let the norm document consist of $n$ sequential sentences $\{x_1, x_2, ..., x_n\}$. Each sentence $x_i$ is encoded using a pre-trained sentence transformer, $s_i \in \mathbb{R}^d$. We define the pairwise cosine similarity matrix $\mathbf{S} \in \mathbb{R}^{n \times n}$:

\begin{equation}
    \mathbf{S}_{i,j} = \frac{s_i \cdot s_j}{\|s_i\|\|s_j\|}
\end{equation}

When visualized as a heatmap, $\mathbf{S}$ reveals a clear block-diagonal structure, clusters of high similarity corresponding to contiguous segments of norm-related sentences. These blocks emerge naturally because norm definitions and their illustrative examples tend to form coherent segments within the text. To segment these into preliminary clusters, we use a greedy line-wise algorithm. We begin with sentence $x_1$ and append subsequent sentences until the similarity with the previous line falls below a segmentation threshold $\epsilon{_\text{seg}}$. Formally, for each $i$, we assign $x_i$ to the current cluster if
\begin{equation}
    \cos(s_{i-1}, s_i) \geq \epsilon_{\text{seg}}
\end{equation}
Otherwise, a new cluster is initiated. We find that each cluster closely represents a candidate \emph{norm chunk} $C_k$, a semantically consistent unit of norm information. Initial segmentation often leads to over-fragmentation due to outlier examples or edge-case scenarios within the same norm description. To address this, we apply a refinement stage: adjacent clusters $C_k$ and $C_{k+1}$ are merged if the cosine similarity between their mean embeddings $\mu_k$ and $\mu_{k+1}$ exceeds a merging threshold $\epsilon_{\text{merge}}$:
\begin{equation}
    \cos(\mu_k, \mu_{k+1}) > \epsilon_{\text{merge}}
\end{equation}

Note, in experimentation we used grid search to find the optimal value for $\epsilon_{\text{seg}}$ and $\epsilon_{\text{merge}}$.
This two-phase process ensures that each norm chunk captures a semantically cohesive unit while minimizing cross-norm contamination. In practice, we observe that a single norm may be split across two clusters, however, it is rare for a single cluster to span multiple norms, thereby validating the precision of our clustering approach. These discovered norm chunks form the foundational retrieval units for semantic alignment with dialogue, enabling fine-grained context-aware norm reasoning in the subsequent stages of our framework.

\subsection{Structured Norm Attribute Extraction}
Real-world dialogue often evolves dynamically, with shifting intents, roles, and relationship framings. Different from general pipelines, the core insight driving our \textbf{Norm-RAG} is that normative behavior in conversation is multi-dimensional, governed not just by surface-level semantics. We represent these multidimensional structures around four foundational attributes of normative behavior: 

\begin{tcolorbox}[
    colback=gray!10,
    colframe=black,
    boxrule=0.4pt,
    arc=2pt,
    left=4pt,
    right=4pt,
    top=4pt,
    bottom=4pt,
    width=\columnwidth,
    enhanced,
    sharp corners
]
\small
\textbf{1. Communicative Intent (\textit{CI}):} What is the pragmatic goal conveyed (e.g., apologize, warn, inquire)?\\
\textbf{2. Interpersonal Framing (\textit{IF}):} What social relationship or power dynamic is implied (e.g., elder-junior, peer-peer)?\\
\textbf{3. Linguistic Features (\textit{LF}):} Which discourse cues or syntactic patterns characterize the norm (e.g., hedging, imperative verbs)?\\
\textbf{4. Contextual Triggers and Constraints (\textit{CTC}):} What environmental, cultural, or situational conditions activate or constrain the norm?
\end{tcolorbox}
These dimensions capture latent communicative goals, social relationships, stylistic strategies, and situational factors that are essential to identifying and interpreting normative behavior. Next, we describe how these attributes are extracted for both the extracted norm chunks and dialogues during inference.

Once the norm chunks $\{C_k\}$ are identified, we extract their attributes via prompting a LLM, \emph{e.g.,} gpt-4o-mini~\cite{hurst2024gpt}. 
The goal is to extract the four underlying components $(CI, IF, LF, CTC)$ from each norm chunk $C_k$ in a normalized format:
\begin{equation}
C_k \xrightarrow{\mathcal{P}_{\text{GPT}}} N_{C_k}:\{ CI_{C_k}, IF_{C_k}, LF_{C_k}, CTC_{C_k} \}
\end{equation}

The prompt is designed to be culturally sensitive and context-aware, mirroring the structure of the norm examples provided in the documentation. 
The LLM returns a structured JSON-like response with textual content or canonical tags corresponding to each attribute. This extraction process not only supports semantic alignment in downstream retrieval, but also introduces interpretability and modularity in norm representation. By decoupling normative knowledge into interpretable attributes, our method enables precise matching between dialogue utterances and applicable social norms, supporting both generalization and cultural specificity in norm understanding.

\subsection{Attribute-based Semantic Retrieval and Re-ranking}

For a dialogue segment $D_t$ up-to time $t$, our goal is to retrieve semantically aligned norm definitions based on shared pragmatic structure. We begin by extracting a normative attribute vector from $D_t$ using the same four-part schema introduced earlier:
\begin{equation}
    D_t \xrightarrow{\mathcal{P}_{\text{GPT}}} N_{D_t}: \{ CI_{D_t}, IF_{D_t}, LF_{D_t}, CTC_{D_t} \}
\end{equation}

Each attribute $a^i_{D_t}$ is compared independently against its corresponding attribute $a^i_{C_k}$ from all norm chunks in the index. Let $\phi$ denote the cosine similarity between embeddings. We define the aggregated similarity score between $N_{D_t}$ and $N_{C_k}$ as:
\begin{equation}
    \text{sim}_k = \frac{1}{4} \sum_{i=1}^{4} \phi(a^i_{D_t}, a^i_{C_k})
\end{equation}

We retrieve top-$n$ candidate norm chunks with the highest $\text{sim}_k$ values, filtered to exceed a global threshold $\mu_{\text{sim}}$ (average across all $\text{sim}_k$ scores), to ensure semantic proximity. These candidates serve as context for grounding downstream prompt generation.

\paragraph{Dialogue Window Design:} 
To prevent topic drift from full-dialogue embeddings, we adopt a recency-focused strategy. Let $u_t$ be the utterance at time $t$ and $D_t$ the entire dialogue segment up till time $t$. We construct a focused context window:
\begin{equation}
    \bar{D_t} = \text{Concat}(u_{t-l}, \dots, u_t)
\end{equation}

This structure only considers the current utterance and previous $l$ dialogue turns, thereby helping preserve critical dependencies while emphasizing recency. Here, $l$ is determined via LLM given the dialogue history and the latest utterance. We optionally apply weighted averaging of embeddings with positional bias toward $u_t$ to capture moment-level intent without any prior contextual noise.

\paragraph{Re-ranking Module:}
The retrieved candidates are further evaluated using a LLM-based re-ranker. Given extracted query attributes $\{a^i_{D_t}\}$ and each norm chunk $N_n$, the reranker $\mathcal{R}$ evaluates contextual alignment and re-ranks the retrieved norm candidates.
The top-ranked norm is accompanied by a natural language explanation generated by the reranker, which justifies its contextual relevance to the dialogue. This attribute-disentangled retrieval mechanism mirrors the interpretive reasoning humans apply when aligning utterances with social norms. It also provides a semantically grounded context that enhances prompt-based inference in the final classification stage.

\subsection{Dialogue-Aware Norm Detection Agent}
In the final stage of the \textbf{Norm-RAG} pipeline, our objective is to identify at each time step $t$, the relevant social norm category and status for each turn within the dialogue session. Additionally, we are also interested in generating explanations for the predicted result and relevant feedback to help at the next turn in the dialogue. To do this our RAG-based agentic framework incrementally processes each dialogue turn, one utterance at a time. At step $t$, the agent receives the focused context window $\bar{D}_t$; re-ranked retrieved norm chunks $\{N^*_1, \dots, N^*_n\}$; and dialogue attributes $N_{D_t}: \{ CI_{D_t}, IF_{D_t}, LF_{D_t}, CTC_{D_t} \}$. These along with the dynamic feedback variable $F_{t-1}$ from the previous turn is embedded as a structured prompt and presented to the LLM for classification. Note, the feedback encodes high-level observations (e.g., tone, contradiction, escalation) that can influence future norm adherence. For the first utterance ($t=1$), we set $F_0 = \emptyset$.




\noindent \textbf{LLM Query and Output:}
The different LLM query inputs and outputs are illustrated as follows: 
\begin{tcolorbox}[title=LLM Inference at Each Step, boxsep=1pt, top=1pt, bottom=1pt, left=2pt, right=2pt]
At each utterance step $t$, the LLM is prompted with:
\begin{itemize}\setlength{\itemsep}{0.5pt}
    \item Focused Dialogue History: $\bar{D}_t$
    \item Retrieved Chunk Attributes: $CI_{D_t}, IF_{D_t},$\\
\hspace*{0.2em}$LF_{D_t}, CTC_{D_t}$
    \item Retrieved Norm Context: $\{N^*_1, \dots, N^*_n\}$
    \item Prior Feedback: $F_{t-1}$
\end{itemize}
The model produces:
\begin{itemize}\setlength{\itemsep}{0.5pt}
    \item Norm Category: $y_t$
    \item Norm Status: $s_t \in \{\text{adhered}, \text{violated}\}$
    \item Explanation for $y_t$ and $s_t$
    \item Updated Feedback Signal: $F_t$
\end{itemize}
\end{tcolorbox}

Formally, the model performs:
\begin{equation}
\small
(y_t, s_t, F_t) = \mathcal{L}(\bar{D}_t, CI_t, IF_t, LF_t, CTC_t, \{N^*_i\}_{i=1}^k, F_{t-1})
\end{equation}
where $\mathcal{L}$ denotes the LLM invoked with a custom instruction-tuned prompt. This agentic setup enables feedback-driven norm reasoning over the course of an entire conversation.

\noindent \textbf{Agentic Loop:}
This structure forms a dynamic reasoning loop over the session:
\begin{tcolorbox}[title=Agentic Reasoning Loop, boxsep=1pt, top=1pt, bottom=1pt, left=2pt, right=2pt]
For a given session with utterances $\{u_1, \dots, u_T\}$:
\begin{enumerate}[itemsep=0.5pt]
    \item Initialize $F_0 = \emptyset$
    \item For $t = 1$ to $T$:
    \begin{itemize}[leftmargin=0.5em, itemsep=0.5pt]
        \item Extract $\bar{D}_t = \{u_1, ..., u_{t-1}\}$
        \item Retrieved Chunk Attributes: $CI_{D_t}, IF_{D_t},$\\
\hspace*{0.2em}$LF_{D_t}, CTC_{D_t}$
        \item Prompt Construction:  $\bar{D}_t$, $N_{D_t}$, and $F_{t-1}$
        \item Query LLM to obtain $(y_t, s_t, F_t)$
    \end{itemize}
    \item Repeat until end of session
\end{enumerate}
\end{tcolorbox}

By iteratively grounding each utterance in structured context, retrieved norms, and conversational dynamics, the \textbf{Norm-RAG} agent enables robust and explainable norm classification throughout multi-turn interactions. This agentic structure also supports temporal coherence, capturing how normative behavior evolves within a session.


\section{Experiments}\label{section_experiments}

We conduct a comprehensive evaluation of our proposed approach on norm discovery and adherence/violation classification for conversations that include multi-lingual and cross-culture scenarios.  

There has been no baselines or proposed model for Norm Discovery and Adherence discovery excluding  NormSage~\cite{fung2022normsage} and NormDial~\cite{li2023normdial}. However, these works do not compare to our proposed dataset and metric. Hence, we include the latest state-of-the-art LLMs both proprietary (GPT-4o ~\cite{hurst2024gpt}) and open-sourced (LLaMA~\cite{dubey2024llama}, QWEN~\cite{yang2025qwen3}, Phi~\cite{abdin2024phi}) baselines for comparison.
For each utterance 



\subsection{Norm Classification and Adherence Detection Accuracy}

Given a conversation, the task is to classify which social norm categories are invoked by the \emph{latest utterance}, and determine whether the utterance adheres to or violates them. We evaluate this across two subtasks: norm classification and adherence status detection.

We benchmark our \textbf{Norm-RAG} framework against a range of zero-shot LLM baselines both closed-source (GPT-4o-mini) and open-source (LLaMA 3.1 8B, Qwen-3 32B, Phi-4 14B), across two key input configurations: 1. \textbf{Hist.}: Whether the model is provided with only the \emph{latest utterance} or with the \emph{entire conversation history + latest utterance}, 2. \textbf{Docs.}: Whether the model receives \emph{no external context} or is given \emph{retrieved documentation} containing culture-specific norm definitions and examples. Each baseline is evaluated under these combinations, while our \textbf{Norm-RAG} method additionally integrates context i.e., structured attributes and feedback-driven reasoning.

As shown in \Cref{tab:combined_results} and \Cref{tab:combined_results_socialdial}, our approach consistently outperforms all baseline configurations across both norm classification and adherence detection tasks. Within each model family, the addition of historical context and external normative knowledge leads to meaningful gains. \textbf{Norm-RAG} further amplifies this by grounding LLM inference in retrieved, semantically-aligned norm definitions. \textbf{Norm-RAG} achieves an improvement of $+14.1\%$, $+21.7$ improvement for LLaMA 3.1 in norm classification and adherence detection accuracy respectively and lowest for  Phi-4 with $~+4.\%$ in norm classification while a drop in $~-2\%$ in adherence accuracy. The modest gains with Phi-4 suggest that its performance benefits less from additional context, likely due to its reliance on internal knowledge and preference for concise instructions—larger retrieved prompts may disrupt its reasoning compared to models like Qwen and LLaMA that better utilize external context. Overall our results confirms that retrieval-augmented structure-aware prompting yields stronger generalization across dialog settings.

\begin{table}[t]
\centering
\small
\setlength{\tabcolsep}{5pt}
\begin{tabularx}{\columnwidth}{
    l
    >{\centering\arraybackslash}p{0.65cm} 
    >{\centering\arraybackslash}p{0.65cm} 
    >{\centering\arraybackslash}p{1.25cm} 
    >{\centering\arraybackslash}p{1.25cm}
}
\toprule
\textbf{Model} & \textbf{Hist.} & \textbf{Docs} & \textbf{Norm Acc.} & \textbf{Adh./Viol. Acc.} \\
\midrule
GPT-4o-mini & \xmark & \xmark & $64.6$ & $57.2$ \\
\textbf{Ours w/ GPT-4o} & \cmark & \cmark & $\textbf{70.4}$ & $\textbf{63.6}$ \\
\addlinespace
LLaMA 3.1 (8B) & \xmark & \xmark & $56.4$ & $44.7$ \\
LLaMA 3.1 (8B) & \cmark & \xmark & $57.4$ & $45.4$ \\
\textbf{Ours w/ LLaMA} & \cmark & \cmark & $\textbf{64.4}$ & $\textbf{54.4}$ \\
\addlinespace
Qwen-3 (32B) & \xmark & \xmark & $61.1$ & $54.9$ \\
Qwen-3 (32B) & \cmark & \xmark & $62.3$ & $57.1$ \\
\textbf{Ours w/ Qwen} & \cmark & \cmark & $\textbf{67.9}$ & $\textbf{60.2}$ \\
\addlinespace
Phi-4 (14B) & \xmark & \xmark & $66.4$ & $58.6$ \\
Phi-4 (14B) & \cmark & \xmark & $67.8$ & $\textbf{61.6}$ \\
\textbf{Ours w/ Phi} & \cmark & \cmark & $\textbf{69.1}$ & $60.2$ \\
\bottomrule
\end{tabularx}
\caption{
\small
Accuracy for norm classification and adherence/violation detection on the MINDS corpus. \textbf{Hist.}: dialogue history used (\cmark) vs.\ only last utterance (\xmark). \textbf{Docs}: retrieved documentation used (\cmark) or not (\xmark).
}
\label{tab:combined_results}
\end{table}

\begin{table}[t]
\centering
\setlength{\tabcolsep}{5pt}
\begin{tabularx}{\columnwidth}{
    l
    >{\centering\arraybackslash}p{1.0cm} 
    >{\centering\arraybackslash}p{1.0cm} 
    >{\centering\arraybackslash}p{1.5cm} 
}
\toprule
\textbf{Model} & \textbf{Hist.} & \textbf{Docs} & \textbf{Norm Acc.} \\
\midrule
GPT-4o-mini & \cmark & \xmark & $58.0$ \\
\textbf{Ours w/ GPT-4o} & \cmark & \cmark & $\textbf{62.0}$ \\
\addlinespace
LLaMA 3.1 (8B) & \cmark & \xmark & $57.2$  \\
\textbf{Ours w/ LLaMA} & \cmark & \cmark & $\textbf{62.0}$  \\
\addlinespace
Qwen-3 (32B) & \cmark & \xmark & $52.4$  \\
\textbf{Ours w/ Qwen} & \cmark & \cmark & $\textbf{55.0}$  \\
\bottomrule
\end{tabularx}
\caption{
\small
Accuracy for norm classification and adherence/violation detection on the SocialDial corpus. \textbf{Hist.}: dialogue history used (\cmark) vs.\ only last utterance (\xmark). \textbf{Docs}: retrieved documentation used (\cmark) or not (\xmark).
}
\label{tab:combined_results_socialdial}
\end{table}

\subsection{Ablation Study}
To better understand the contribution of different components in the \textbf{Norm-RAG} framework, we conduct an ablation study summarized in \Cref{tab:ablation}. Starting with a fixed-size text splitter chunking mechanism (row 1), we observe significantly lower performance across both models, suggesting that arbitrary chunking limits semantic cohesion in the retrieved context. When switching to semantically clustered chunks without extracting norm attributes (row 2), we see consistent improvements (+1.9 in Qwen and +1.3 in GPT for norm accuracy), indicating the value of topically coherent segmentation. Turning off feedback (row 3) yields additional gains, showing that conversational grounding plays a role in cumulative norm reasoning. Finally, the full \textbf{Norm-RAG} pipeline (row 4) achieves the best performance across both models, confirming that attribute-guided context and feedback-driven prompting are complementary. Notably, GPT shows a larger gain from feedback than Qwen, likely due to its stronger adaptation to turn-level dialog intent modeling.

\begin{table}[t]
\centering
\setlength{\tabcolsep}{5pt}
\begin{tabularx}{\linewidth}{
    l
    >{\centering\arraybackslash}p{2cm} 
    >{\centering\arraybackslash}p{2cm}
}
\toprule
\textbf{Ablation Variant} & \textbf{Qwen-3} & \textbf{GPT-4o} \\
\midrule
Fixed text split & $63.2$ / $58.7$ & $66.1$ / $61.2$ \\
Cluster w/o attr. & $65.1$ / $59.0$ & $67.4$ / $61.9$ \\
w/o feedback & $67.0$ / $60.1$ & $69.2$ / $62.9$ \\
Full pipeline & $\textbf{67.9}$ / $\textbf{60.2}$ & $\textbf{70.4}$ / $\textbf{63.6}$ \\
\bottomrule
\end{tabularx}
\caption{
\small
Ablation study of the \textbf{Norm-RAG} pipeline. "Fixed text split" uses uniform segmentation; "Cluster w/o attr." excludes attribute extraction; "w/o feedback" disables conversational feedback. We report norm classification and status accuracy as Norm/Status.
}
\label{tab:ablation}
\end{table}

\subsection{Qualitative Analysis}
In challenging cases involving overlapping intents, such as: ``I want the junior surgeon who performed the operation to be fired." \textbf{Norm-RAG} correctly identifies the \texttt{doing request} as an act of adherence, while also detecting the embedded \texttt{doing criticism} as a norm violation—matching the human-labeled ground truth. By leveraging structured dialog attributes and semantically retrieved normative context, Norm-RAG disentangles multi-intent utterances and reasons about their distinct normative implications with high fidelity, even when cues are emotionally charged or culturally sensitive.





\section{Conclusion}\label{section_conclusion}
We presented \textbf{Norm-RAG}, a retrieval-augmented, agentic framework for modeling social norm adherence in multilingual dialogue. By combining semantically clustered norm documentation with structured dialog attributes—such as communicative intent, framing, and linguistic features—our approach enables interpretable and culturally sensitive norm reasoning. We introduced \textbf{MINDS}, a novel bilingual dataset of Mandarin-English and Spanish-English conversations annotated for norm categories and adherence status. Empirical results across multiple LLMs demonstrate significant gains in both norm classification and adherence detection. Through ablation studies, we show the importance of semantic chunking, structured attribute modeling, and feedback-based prompting. Together, these components move beyond static, template-based inference toward dynamic, socially grounded interaction. Future work will explore higher-order norm dynamics such as escalation, social repair, and longitudinal alignment in multilingual, multi-agent settings.

\section*{Acknowledgments}
This material is based upon work supported by the Defense Advanced Research Projects Agency
(DARPA) under Contract No. HR001122C0032. Any opinions, findings and conclusions or recommendations expressed in this material are those of the author(s) and do not necessarily reflect the
views or policies of DARPA, the Department of Defense or the U.S. Government. Additionally, we acknowledge the use of GPT-based language models during the development of this work for research assistance and code prototyping. These tools supported various stages of exploration, analysis, and implementation.

\newpage
\section*{Limitations and Ethical Considerations}
While our work aims to improve norm understanding in multilingual dialogues, it has several limitations. First, our dataset covers only two non-English languages (Mandarin and Spanish), which limits generalization to other linguistic and cultural contexts. Second, although we employ LLMs for attribute extraction and classification, these models can inherit social and cultural biases from their training data, which may affect norm interpretation. We also acknowledge the limitations of using simulated or partially controlled bilingual dialogues, which may differ from spontaneous in-the-wild interactions.

Ethically, we ensured speaker anonymization in our dataset and limited language use to non-sensitive topics. We encourage future work to expand linguistic coverage and explore real-world deployment risks in downstream applications.

\paragraph{Computational Resources}
We conducted all experiments using 2× NVIDIA A6000 GPUs (each with 48GB VRAM). The classification pipeline used three open-source LLMs: LLaMA 3.1 (8B), Qwen (32B), and Phi-3 (14B), applied to a dataset of 835 utterances. For each utterance, the agent performed structured prompting and reranking per model. The total compute time across all models was approximately ~ 30 GPU hours on the A6000 infrastructure.

\bibliography{custom}

\appendix
\newpage

\section{Prompts Used in \textbf{Norm-RAG}}\label{sec:prompt}

\lstdefinestyle{wrappedcode}{
  language=Python,
  basicstyle=\ttfamily\footnotesize,
  breaklines=true,
  breakatwhitespace=true,
  breakindent=0pt,
  showstringspaces=false,
  tabsize=2
}

\begin{figure}[ht!]
\centering
\begin{tcolorbox}[title=Norm Attribute Extraction Prompt,
                  colback=lightgray,
                  colframe=black!75!white,
                  sharp corners=southwest,
                  fonttitle=\bfseries,
                  width=\textwidth]
\textbf{System Prompt}
\begin{lstlisting}[style=wrappedcode]
"""You are a social interaction analyst specializing in pragmatics and social norm recognition in conversation.

Given the dialogue below, extract a structured representation of the **speaker's behavior in the final utterance**, focusing on how it performs or aligns with one or more socially recognizable norms such as persuasion, request, refusal, apology, etc.

The extracted structure will be used to retrieve similar conversational behaviors, so it must **accurately reflect the speech act's social function, nuance, and framing**, in a way that can **disambiguate between multiple norm categories**.

=> If the utterance aligns with **more than one norm** (e.g., 'doing request' + 'doing thanks'), your attributes should reflect that layered action.

Return the following 4 **pragmatic attributes**:

```json
{{
  ``CommunicativeIntent": ``<Describe *all communicative goals* the speaker is pursuing -- both primary and secondary. Use norm language if applicable (e.g., persuading, requesting info, refusing, finalizing). Prioritize intent differentiation across norms.>",
  ``InterpersonalFraming": ``<How the speaker *relates to the listener*: formality, power dynamics, face-work (saving/threatening), emotional stance, or alignment. Make distinctions like deferential vs. assertive, affiliative vs. distancing -- as they cue norm categories.>",
  ``LinguisticFeatures": ``<Detail rhetorical strategies used to *signal or mitigate norm performance*: hedges, indirectness, modality (e.g., 'might', 'should'), discourse markers, politeness formulas, etc. Capture evidence that helps distinguish one norm from another.>",
  ``ContextualTriggersAndConstraints": ``<What about the broader dialogue or situation shapes how this norm is performed? Include role relations, timing, known stakes, prior acts, social rules or expectations that constrain the speaker's behavior.>"
}}
```
"""
\end{lstlisting}
\vspace{1em}
\textbf{User Prompt}
\begin{lstlisting}[style=wrappedcode]
### Dialogue Context:
{dialog_context}
\end{lstlisting}

\end{tcolorbox}
\label{fig:prompt_norm_attribute_extraction}
\end{figure}

\begin{figure*}[t]
\centering
\begin{tcolorbox}[title=Dialog Window Design Prompt,
                  colback=lightgray,
                  colframe=black!75!white,
                  sharp corners=southwest,
                  fonttitle=\bfseries,
                  width=\textwidth]

\textbf{System Prompt}
\begin{lstlisting}[style=wrappedcode]
You are an expert in pragmatics and social norms.
Given the dialogue history below, analyze the communicative function and social dynamic of the most recent turn.
Please return your response in the following JSON format:
```
{{
    "CommunicativeIntent": "<short summary of what the speaker is trying to achieve>",
    "InterpersonalTension": "<comment on any social tension, repair, dominance, submission, etc.>",
    "LikelyNormCategory": "<the most likely norm involved, e.g., 'doing apology', 'doing greeting', etc.>",
    "ContextDependenceScore": <float between 0.0 and 1.0, where higher means more dependent on prior context>
}}
```
\end{lstlisting}

\vspace{1em}
\textbf{User Prompt}
\begin{lstlisting}[style=wrappedcode]
### Dialogue History:
{dialog_history}
\end{lstlisting}

\end{tcolorbox}
\label{fig:prompt_dialog_window_design}
\end{figure*}

\begin{figure*}[ht!]
\centering
\begin{tcolorbox}[title=Re-ranking Retrieved Norm Chunks Prompt,
                  colback=lightgray,
                  colframe=black!75!white,
                  sharp corners=southwest,
                  fonttitle=\bfseries,
                  width=\textwidth]

\textbf{Prompt}
\begin{lstlisting}[style=wrappedcode]
You are a pragmatics and discourse analysis expert.
You are given:
-- A brief snippet of dialogue (usually the last 1-2 turns of a conversation),
-- A structured interpretation of that snippet, for attribute {attribute_name},
-- A list of candidate norm definitions retrieved from a semantic search system.
Your task is to rerank the candidates from most to least relevant, based on how well each one aligns with the communicative behavior expressed in the dialogue as represented by the extracted attributes.

### Dialogue Context:
"{dialog_context}"

### Extracted Norm Attributes:
{attributes}

### Retrieved Candidate Norm Descriptions:
{doc_entries}

### Instructions:
-- Compare the overall meaning and function of each candidate to the extracted attributes.
-- Pay special attention to the Communicative Intent, but also consider whether the interpersonal stance, language choices, and situational framing match.
-- Your goal is to rank which candidate best captures the type of norm being enacted in the given dialogue.

### Output Format:
{{
  "Ranking": [3, 1, 2, 4, 5],
  "TopJustification": "..."
}}
Only return the JSON object.
\end{lstlisting}

\end{tcolorbox}
\label{fig:prompt_rerank_retrieved_chunks}
\end{figure*}

\begin{figure*}[t]
\centering
\begin{tcolorbox}[title=Feedback Prompt,
                  colback=lightgray,
                  colframe=black!75!white,
                  sharp corners=southwest,
                  fonttitle=\bfseries,
                  width=\textwidth]

\textbf{System Prompt}
\begin{lstlisting}[style=wrappedcode]
You are a pragmatic analyst helping to generate interpretive context for understanding turn-by-turn norms in conversation.
Given the most recent utterance in a dialogue, along with its predicted norm(s) and surrounding dialogue context, your task is to produce **feedback that captures the communicative force and social trajectory** of the current moment.
This feedback will be used to inform the interpretation of the *next* utterance -- by helping identify what norms or responses are socially relevant or expected, and what social constraints are already in play.

### INPUT:
- `DialogueHistory`: The full dialogue history leading up to the latest utterance (short or long).
- `LastUtterance`: The final utterance by the most recent speaker.
- `PredictedNorms`: One or more social norms inferred from the last utterance. One or more of:
  ['Doing persuasion', 'Doing request', 'Doing requesting information', 'Doing criticism', 'Doing thanks', 'Doing greeting', 'Doing admiration', 'Doing disagreement', 'Doing refusing a request', 'Doing apology', 'Doing taking leave', 'Doing granting a request', 'Doing finalizing negotiation/deal', 'No Norm']

### OUTPUT FORMAT:
{
  ``SituatedSummary": ``<Explain what is being socially performed in the last utterance, and how it connects to the unfolding dialogue -- including tone, intentions, relational shifts, or embedded expectations.>",
  ``NormImplications": ``<What social norm(s) are being enacted or invoked? Why? Include cues from wording, context, or sequencing.>",
  ``NextTurnExpectation": ``<What types of responses -- in terms of social action or stance -- are made relevant by this utterance? What does it *invite*, *pressure*, or *allow* the next speaker to do (or not do)? Mention if there's a power dynamic, politeness constraint, emotional charge, etc.>"
}
\end{lstlisting}

\vspace{1em}
\textbf{User Prompt}
\begin{lstlisting}[style=wrappedcode]
### Dialogue History
{dialoghistory}

### Last Utterance
{lastutterance}

### Predicted Norms
{predictednorms}
\end{lstlisting}

\end{tcolorbox}
\label{fig:prompt_feedback}
\end{figure*}

\begin{figure*}[t]
\centering
\begin{tcolorbox}[title=Norm Detection and Adherence Detection Prompt,
                  colback=lightgray,
                  colframe=black!75!white,
                  sharp corners=southwest,
                  fonttitle=\bfseries,
                  width=\textwidth]

\textbf{System Prompt}
\begin{lstlisting}[style=wrappedcode]
You are an expert in analyzing conversations to identify underlying social norms. Your task is to classify all applicable social norm categories (minimum 2, maximum upto 5) reflected in the **latest utterance** of a given dialogue using both **explicit and implicit cues** of social interaction.
### Norm Categories:
{norm_categories}
### Task Instructions:
1. Use the **entire dialogue history** and the **retrieved context from RAG** to interpret the **social intent** behind the **latest utterance**.
   - Consider both **explicit speech acts** (e.g., asking, refusing) and **implicit or indirect signals** (e.g., persuading by justification, criticizing through description).
   - Understand the progression and structure of the dialogue to reveal the **pragmatic function** of the utterance.
2. Identify **all relevant norm categories** the latest utterance satisfies from the list (maximum 3).
   - Choose norms based on **intent**, **emotion**, **relational context**, **dialogue progression**, and **linguistic cues**, even when **indirectly expressed**.
   - Include **weak or moderate instances** of norms (e.g., subtle persuasion or soft disagreement), not just overt ones.
3. For each norm category:
   - Assess whether the utterance reflects **Adherence** or **Violation** of that norm.
4. Evaluate whether the **retriever context** is relevant to the **overall set of predicted norms**:
   - If **Relevant**, use it to support a more confident classification.
   - If **Not Relevant**, ignore the retriever context and use your own reasoning about social norms.
5. If **no identifiable norm** is present in the utterance:
   - Return only one entry:
     - Norm Category: `No Norm`
     - Status: `Violation`
6. Provide a **natural language confidence level** for your prediction:
   - Choose from: `High`, `Medium`, or `Low`
   - Justify your confidence based on clarity of social intent, surface and hidden patterns, and context fit.

### Output Format in JSON:
```json
{{
  "latest_utterance": "<copy of the utterance>",
  "predicted_norms": [
    {{"norm_category": "<norm category 1>", "status": "<Adherence or Violation>"}},
    {{"norm_category": "<norm category 2>", "status": "<Adherence or Violation>"}},
    {{"norm_category": "<norm category 3>", "status": "<Adherence or Violation>"}}
  ],
  "retriever_context_relevance": "<Relevant / Not Relevant>",
  "confidence_level": "<High / Medium / Low>",
  "explanation": "<Justify the norm predictions, referencing how context and implicit cues shaped the interpretation>"
}}
```
\end{lstlisting}
\textbf{User Prompt with (or without) Context}
\begin{lstlisting}[style=wrappedcode]
{% if context %}
### Relevant Context from RAG on 4 key attributes that are used to capture the underlying norm:
{{ context }}
{% endif %}
### Dialog:
{{ dialog }}
\end{lstlisting}

\end{tcolorbox}
\label{fig:prompt_norm_adherence_detection}
\end{figure*}

\end{document}